\documentclass[letterpaper]{article} 
\usepackage{aaai25}  
\usepackage{times}  
\usepackage{helvet}  
\usepackage{courier}  
\usepackage[hyphens]{url}  
\usepackage{graphicx} 
\usepackage{natbib}  
\usepackage{caption} 
\usepackage{algorithm}
\usepackage{algorithmic}
\usepackage{amsmath}
\usepackage{amssymb}
\usepackage{utfsym}
\usepackage{bbding}
\usepackage{booktabs}
\usepackage{multirow}

\usepackage{newfloat}
\usepackage{listings}
\DeclareCaptionStyle{ruled}{labelfont=normalfont,labelsep=colon,strut=off} 
\floatstyle{ruled}
\newfloat{listing}{tb}{lst}{}
\floatname{listing}{Listing}
\title{LDP: Generalizing to Multilingual Visual Information Extraction by Language Decoupled Pretraining}
\author{
    Huawen Shen\textsuperscript{\rm 1,3},
    Gengluo Li\textsuperscript{\rm 1,3},
    Jinwen Zhong\textsuperscript{\rm 1}\thanks{Corresponding authors},
    Yu Zhou\textsuperscript{\rm 2}\footnotemark[1]
}
\affiliations{
    \textsuperscript{\rm 1}Institute of Information Engineering, Chinese Academy of Sciences\\
    \textsuperscript{\rm 2}VCIP \& TMCC \& DISSec, College of Computer Science, Nankai University\\
    \textsuperscript{\rm 3}School of Cyber Security, University of Chinese Academy of Sciences\\

    \{shenhuawen, ligengluo, zhongjinwen\}@iie.ac.cn, yzhou@nankai.edu.cn
}

\usepackage{bibentry}

\begin{document}

\maketitle

\begin{abstract}
Visual Information Extraction (VIE) plays a crucial role in the comprehension of semi-structured documents, and several pre-trained models have been developed to enhance performance. However, most of these works are monolingual (usually English). Due to the extremely unbalanced quantity and quality of pre-training corpora between English and other languages, few works can extend to non-English scenarios. 
In this paper, we conduct systematic experiments to show that vision and layout modality hold invariance among images with different languages. If decoupling language bias from document images, a vision-layout-based model can achieve impressive cross-lingual generalization.
Accordingly, we present a simple but effective multilingual training paradigm \textbf{LDP} (\textbf{L}anguage \textbf{D}ecoupled \textbf{P}re-training) for better utilization of monolingual pre-training data.
Our proposed model \textbf{LDM} (\textbf{L}anguage \textbf{D}ecoupled \textbf{M}odel) is first pre-trained on the language-independent data, where the language knowledge is decoupled by a diffusion model, and then the LDM is fine-tuned on the downstream languages.
Extensive experiments show that the LDM outperformed all SOTA multilingual pre-trained models, and also maintains competitiveness on downstream monolingual/English benchmarks.
\end{abstract}

\section{Introduction}

Images with text, such as scanned documents \cite{DRCC} and street views \cite{BeyongVQA}, are widely used in our daily life \cite{TextSurvey}. Given the intricate layout and vision clues present in these documents, merely detecting \cite{TextDetection} and recognizing \cite{SEED,PIMNet} all text in the images and serializing to a text sequence could result in a substantial loss of information \cite{TextRetrival}.
Therefore, Visual Information Extraction (VIE) has been developed, tasking the model with utilizing multi-modal information (including vision, layout, and text) to extract essential information from a variety of documents.
Inspired by advancements in the pre-training fine-tuning paradigm \cite{T5}, numerous studies have been undertaken to further advance this field. Similar to other tasks, VIE also encounters a significant challenge: English corpus dominates the pre-training corpus while other languages lack adequate training.

Most multilingual works opt to collect more non-English pre-training data, which can be either synthetic \cite{Donut} or real-life scenarios \cite{layoutxlm}. However, synthetic data often exhibit template-based structures and lack meaningful sequences, constraining the effectiveness of these approaches. Gathering real data can be time-consuming and costly, and these data are usually limited to certain languages \cite{StrucTexTv2}. 
In this context, LiLT \cite{LiLT} initially focuses on pre-training using available monolingual data and then smoothly transitions to multilingual benchmarks.
In pursuit of this objective, LiLT overlooks the visual modality and only decouples the layout information.
A pertinent query arises: \textbf{\textit{Does the visual modality offer benefits in multilingual VIE? Is it possible to leverage the visual modality to enhance the pre-training of our multilingual model?}}

\begin{figure}[t]
\centering
\includegraphics[width=\columnwidth]{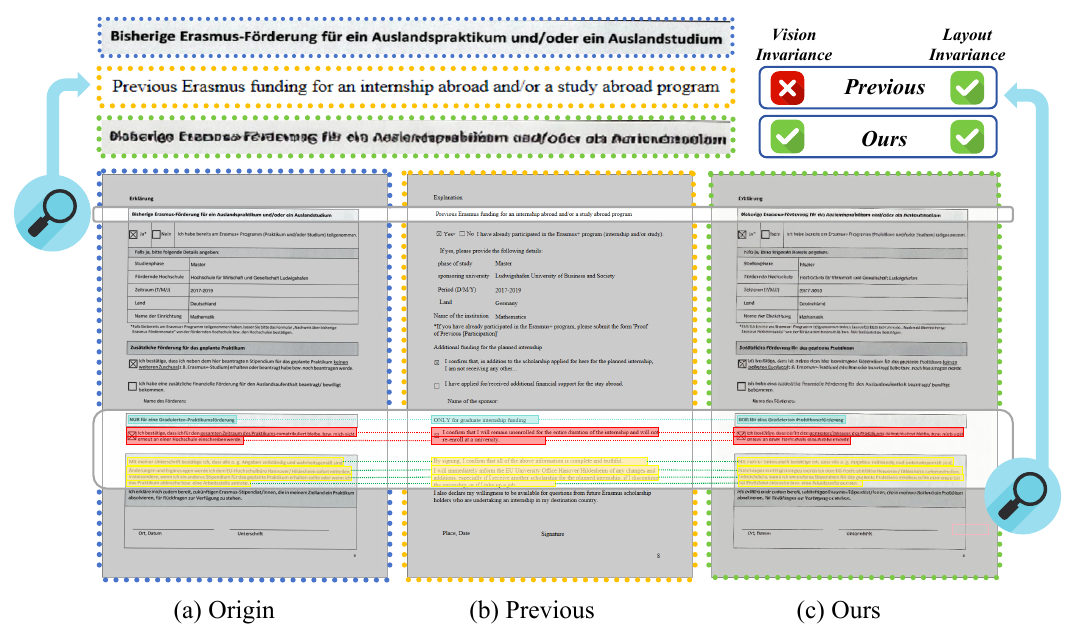}
\caption{(a) Original image. (b) The previous method, LiLT, only decouples the layout modality across different languages, ignoring the vital appearance. (c) Our method remains vision and layout consistent with original image.}
\label{fig:title}
\vspace{-12pt}
\end{figure}

Based on our early attempts, we propose that similar to the layout, visual features exhibit a comparable level of invariance across various languages.
For instance, as shown in Figure \ref{fig:title}(a), though not understanding German words, we can infer that a text sequence in bold font with a distinct gray background typically indicates a section title.
In contrast, when vision information is completely absent, such as Figure \ref{fig:title}(b), the final performance will largely depend on the language model's multilingual capabilities, which might be inconsistent across different languages.

However, images inherently contain language or text information. Even without explicitly training the model to extract this information, it would naturally overfit the target language \cite{StrucTexTv2,TRIE}, limiting its generalization to unseen languages.
To address this, we propose to decouple the language bias from the document images using the text edit diffusion model \cite{AnyText}.
As shown in Figure \ref{fig:title}(a) and (c), the text in the language-decoupled image maintains the original layout and visual features, but it is not associated with any recognizable language. In our experiments, the integration of decoupled data eventually enhances the generalization of unseen languages.

Motivated by this observation, we propose a novel multilingual training paradigm, referred to as \textbf{LDP} (\textbf{L}anguage \textbf{D}ecoupled \textbf{P}re-training), which utilizes only the open-sourced English corpus for pre-training.
For each image in the pre-training stage, we first generate pseudo labels following ESP \cite{ESP}, then employ AnyText to decouple all language bias from the original images. The pseudo labels and language-independent images are used to pre-train our model. Finally, we apply the pre-trained model to fine-tune and test on downstream benchmarks.
The LDP paradigm primarily focuses on addressing the language imbalance in pre-training data volume, where the English corpus plays a dominant role. Language-decoupled data can significantly enhance non-English performance while only slightly reducing English accuracy. In downstream datasets such as XFUND \cite{XFUND}, where the distribution of different languages is balanced, there is no need to decouple language bias, and therefore, the original images are utilized.

To fully utilize the language-independent training data generated by LDP, we propose a simple but effective LDM for information extraction from multilingual document images.
The LDM inherits the SAM (Segment Anything Model) \cite{SAM} framework while replacing SAM's mask prediction head with a randomly initialized MLP head to better suit the VIE task. We follow SAM's preprocessing and encoding procedure. 
However, SAM's mask decoder separately processes different bounding boxes, ignoring the interaction among them. To address this limitation, we introduce the \textbf{MTIM} (Multi-Token Information Merging) module to consolidate information from various bounding boxes within a single image. The enhanced model undergoes pre-training on language-independent data.
In the fine-tuning stage, we introduce the \textbf{LKI} (Language Knowledge Inserting) module to incorporate the decoupled language information into downstream tasks. This integration of language information can significantly enhance the model's performance, particularly in challenging scenarios.

Extensive experimental results demonstrate that the LDM attains state-of-the-art performance on multilingual benchmarks such as XFUND and SIBR, while also preserving comparable monolingual (English) performance when compared to other English-specific models. The primary contributions of our research can be outlined as follows:

\begin{itemize}
\item We are the first to systematically study the visual invariance in the multilingual VIE task. Our findings suggest that decoupling language bias from training data can improve multilingual generalization.
\item We introduce a new language-independent training diagram, LDP, based on our research, which enables the model to generalize across multiple languages using only monolingual pre-training data.
\item Our proposed method, the LDM, achieves state-of-the-art performance in multilingual scenarios while maintaining competitive results in monolingual datasets.
\end{itemize}

\section{Related Work}

\noindent\textbf{Visual Document Pre-training models.} Following the pre-training fine-tuning paradigm in NLP \cite{BERT}, LayoutLM \cite{LayoutLM} first tries to integrate layout with text and applies unsupervised pre-training on a huge amount of document corpus \cite{IIT-CDIP}, achieving impressive results on visual document understanding. LayoutLMv2 \cite{LayoutLMv2} and LayoutLMv3 \cite{LayoutLMv3} further jointly embed vision modality into the model input when in pre-training.
Some works put attention on better organizing the text modality input.
StrucTexT \cite{StrucTexT} applies segment-level embedding to model cross-granularity information, XYLayoutLM \cite{XYLayoutLM} propose a novel XY Cut algorithm to heuristic divide and conquer text to organize a proper reading order. 
Some works also try to maintain an OCR-free paradigm to get better performance in real scenarios where text detection and recognition are not conducted. 
TRIE++ \cite{TRIE++} proposes to jointly train text reading and information extraction in a unified network, StrucTexTv2 \cite{StrucTexTv2} directly performs masked visual-textual prediction in pre-training to get an OCR-free pre-trained model. The generative manner also attracts the attention of the academic community due to its flexible forms. UDOP \cite{UDOP} models several pre-training targets in a generative manner in one framework, DocFormerv2 \cite{DocFormerv2} jointly applies mask pre-training on the encoder and next token prediction on the decoder, outperforming previous work in several downstream tasks. Inspired by the bloom of Large Language Model (LLM) \cite{LLaMA} and Vision Large Language Model (VLLM) \cite{LLaVA}, large models with numerous pre-training data are also proposed. MPLUG-DocOwl \cite{mPLUG-DocOwl} finetune pre-trained mPLUG-Owl \cite{mPLUG-Owl} with document data. LayoutLLM \cite{LayoutLLM} combines pre-trained LayoutLMv3 with LLM to enable LLM better visual document perception. TextMonkey \cite{textmonkey} and mPLUG-DocOwl 1.5 \cite{docowl-1.5} cut the high-resolution document image into several patches to enable large model higher resolution and detailed input.

\noindent\textbf{Multi-lingual models.} Most previous works have mainly focused on English documents, overlooking other languages. XFUND \cite{XFUND} is the first to raise this issue, and they manually label seven non-English datasets with the format same as FUNSD \cite{FUNSD}, making it possible for the industry to examine different models' multi-lingual performance. LayoutXLM \cite{layoutxlm} simply applies the same architecture as LayoutLMv2, and collects numerous multi-lingual pre-training data to re-pre-train the model in the multi-lingual settings. Donut \cite{Donut} generates synthetic multilingual documents using ImageNet \cite{ImageNet} and Wikipedia, applying an auto-regressive generative manner and taking the text reading as the pre-training task. StrucTexTv2 \cite{StrucTexTv2} is further pre-trained on the private Chinese document images to enable the Chinese ability. In the VLLM era, Vary \cite{Vary} applies the auto-regressive text reading task on both English and Chinese data. LiLT \cite{LiLT} first tries to decouple the text modality and layout modality into two branches, and only the layout branch will be optimized in pre-training, which makes different languages share similar text embedding. ESP \cite{ESP} follows a similar manner to TRIE \cite{TRIE}, it takes vision modality as the only input and is only pre-trained in English. Interestingly, ESP can achieve multi-lingual VIE in the downstream dataset, however, no further study has been applied about why ESP obtains the multi-lingual ability.

\section{Are Vision Models Multi-Lingual?}

\subsection{Decoupling Language Bias from Images}

Some previous studies \cite{ESP,TRIE,StrucTexTv2} have used purely visual inputs for (vision-)language training tasks like Mask Language Modeling (MLM) and Image-Text Matching (ITM), indicating that vision-input model can directly acknowledge the language information.
We propose transforming the original document images into a fictional language that does not exist in the real world while retaining the key visual features, such as color, font, and background to avoid introducing language bias into the model, as shown in Figure \ref{fig:title}(a) and (c).

Diffusion-based models often utilize a pre-trained VAE \cite{VAE} tokenizer to encode the entire image. 
Previous works \cite{TextCtrl,VAE-Text2} propose this approach could overlook fine details and distort small objects.
AnyText \cite{AnyText}, designed for conditional text editing, is trained to modify the target region based on the prompt while keeping all other areas unchanged.
However, AnyText is trained on natural scenes, where text is usually large and sparse, unlike the small and dense text found in document images. Our experiments confirmed that it distorts small text in document images, consistent with previous findings by \citeauthor{VAE-Text1}
Motivated by this issue, we employ AnyText to modify our data.

We first resize the original image to a specific resolution, referred as \textbf{``decouple resolution"}. AnyText retains the input image size and applies a fixed-size patch, which leads to more detail loss at smaller resolutions. Thus, ``decouple resolution" can be used as a hyper-parameter to control language bias decoupling.
Next, we specify the pixel in the upper left corner $[0, 0, 1, 1]$ as the target region and add a simple prompt ``\_", where \_ represents a blank space in AnyText’s language tokenizer. Ideally, this instructs the AnyText model to edit the single pixel into blank space, leaving all other regions unedited. But as previously mentioned, the dense text outside the target region will be distorted.
Finally, the distorted images are resized to the original size.

\subsection{Quantitative Evaluation}

\begin{figure}[t]
\centering
\includegraphics[width=0.9\columnwidth]{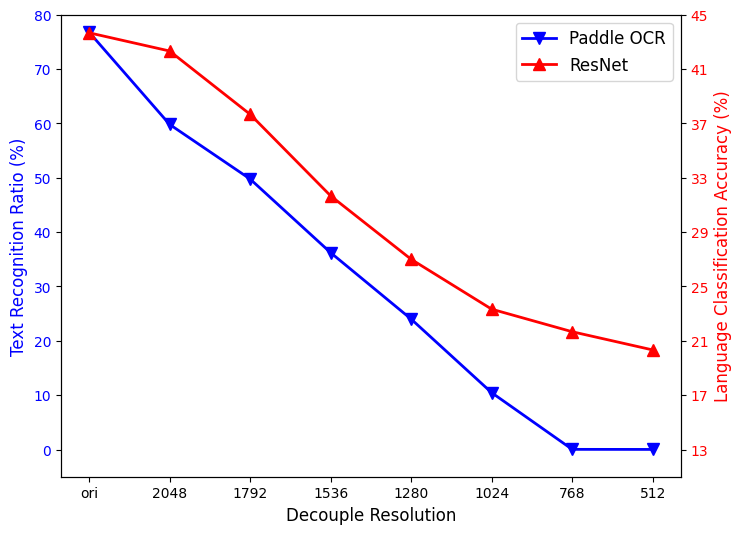}
\caption{The text recognition ratio and language classification accuracy on XFUND. ``ori" means the original image where the language bias is not decoupled by AnyText.}
\label{fig: decoupling}
\end{figure}

\begin{figure}[t]
\centering
\includegraphics[width=0.9\columnwidth]{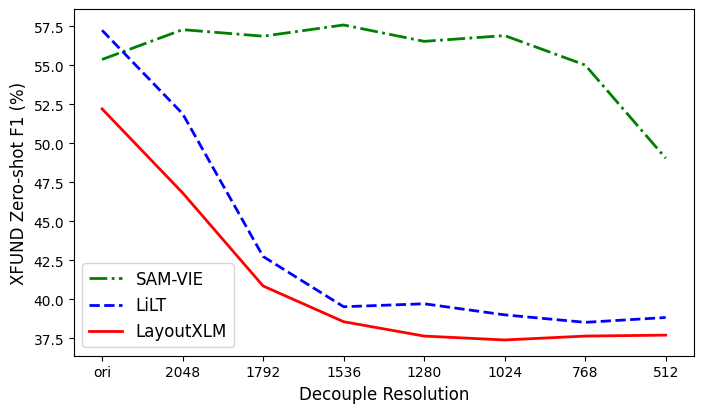}
\caption{VIE performance on XFUND when applying language-decoupled images.}
\label{fig: vie-decoupling}
\vspace{-12pt}
\end{figure}

\begin{figure*}[t]
\centering
\includegraphics[width=1.9\columnwidth]{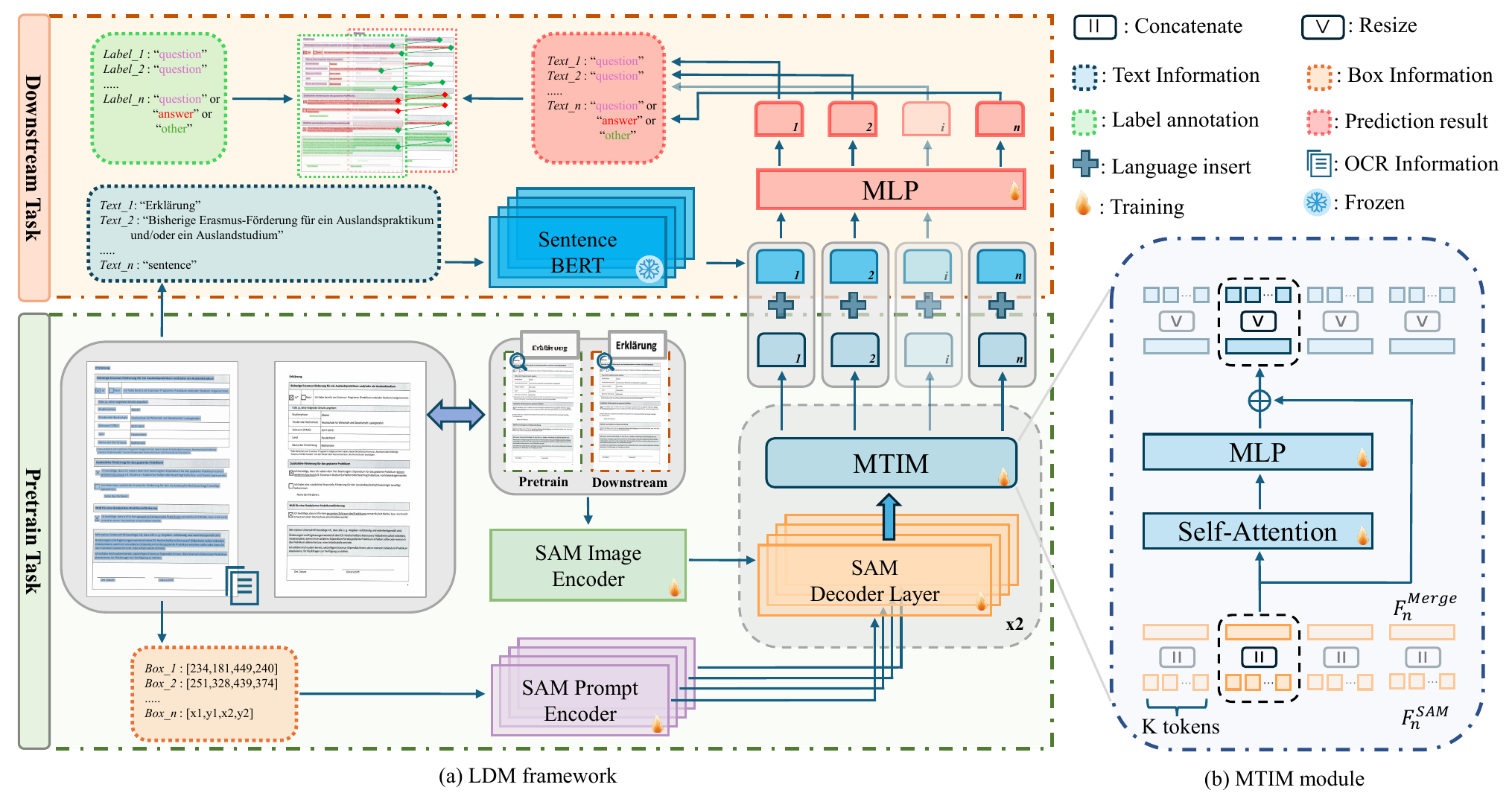}
\caption{The overall illustration of LDM. LDM takes the image and bounding boxes as input, which exactly follows SAM's preprocessing and encoding. After each SAM decoder layer, MTIM is proposed to integrate information from different bounding boxes. A pre-trained frozen Sentence BERT is applied to augment language knowledge for downstream tasks.}
\label{fig: architecture}
\vspace{-12pt}
\end{figure*}

\noindent\textbf{AnyText model can decouple language bias from document images.} To measure the remaining language bias in decoupled images, we conduct experiments on XFUND and use two surrogate metrics: (i) text recognition ratio, and (ii) language classification accuracy. For the text recognition ratio, we crop the image according to the bounding box annotation and apply the PaddleOCR to recognize the text in XFUND test set images. The ratio is calculated as: 

$$
Ratio = 1 - \frac{EditDistance(pred, gt)}{Length(pred) + Length(gt)}
$$

\noindent This metric reflects the extent to which detailed text information is retained. Lower OCR accuracy indicates that it is more challenging for the model to extract text information from purely visual input.
For language classification accuracy, images in a particular XFUND language subset are treated as belonging to the same category. A ResNet model is trained on the XFUND training set and tested on the XFUND test set. Due to the significant visual differences between Chinese and other languages, the Chinese subset is excluded. The more likely two images are classified into the same language category, the less language bias remains in the images.
Results are shown in Figure \ref{fig: decoupling}. As decoupling resolution decreases, both metrics decline correspondingly.
The text recognition ratio decreases almost linearly until the ``decouple resolution" reaches 768, at which point the metric drops to nearly 0 (3.48\%).
For language classification accuracy, the theory lower-bound is 16.66\%, where an image is randomly classified into one of the six languages. The metric approaches this theoretical value, reaching 21.67\% at ``decouple resolution" 768 and 20.33\% at ``decouple resolution" 512. Considering that XFUND images in different languages originate from various sources, there may be language-specific features such as unique form structures or dominant colors. This result suggests that it is nearly impossible for the model to distinguish different languages.

\noindent\textbf{Language-decoupled images can enhance cross-lingual generalization.} 
We evaluate the VIE performance of various models. For language-decoupled images, the text is obtained from PaddleOCR recognition results. All models are fine-tuned on FUNSD (English) and zero-shot evaluated on XFUND (non-English).
LiLT relies entirely on layout and text modalities. LayoutXLM also utilizes visual modalities, but the image resolution is relatively low ($224 \times 224$), making it difficult to capture detailed information.
We also modified the SAM (Segment Anything Model) to serve as the baseline for vision-based models. SAM is pre-trained on instance segmentation tasks with a large number of natural images, making it well-trained with both vision and layout inputs. We simply removed its mask prediction head and replaced it with an MLP layer to predict the label of specific text bounding boxes. Results are shown in Figure \ref{fig: vie-decoupling}.
As language bias is gradually decoupled from the image, both LiLT and LayoutXLM exhibit a significant decrease in performance. For the vision-based model, we observe a slight increase in cross-lingual generalization performance, even surpassing that of well-designed SOTA models.
These observations confirm our hypothesis that decoupling language bias contributes to improved cross-lingual generalization.

\begin{table*}[]
\centering
\begin{tabular}{lcccccccccc}
\toprule
\multirow{2}{*}{Model} & FUNSD & \multicolumn{7}{c}{XFUND}                             & \multicolumn{2}{c}{Avg.}  \\ \cmidrule(l){2-2} \cmidrule(l){3-9} \cmidrule(l){10-11}
                       & EN    & ZH    & JA    & ES    & FR    & IT    & DE    & PT    & All   & w/o EN                  \\ \midrule
XLM-RoBERTa            & 66.70 & 41.44 & 30.23 & 30.55 & 37.10 & 27.67 & 32.86 & 39.36 & 38.24 & 34.17               \\
InfoXLM                & 68.52 & 44.08 & 36.03 & 31.02 & 40.21 & 28.80 & 35.87 & 45.02 & 41.19 & 37.29               \\ \midrule
LayoutXLM              & 79.40 & 60.19 & 47.15 & 45.65 & 57.57 & 48.46 & 52.52 & 53.90 & 55.61 & 52.21              \\
LiLT                   & 84.15 & 61.52 & 51.84 & 51.01 & 59.23 & 53.71 & 60.13 & 63.25 & 60.61 & 57.24             \\ \midrule
\textbf{LDM(Ours)}                    & \textbf{88.23} & \textbf{66.09} & \textbf{57.84} & \textbf{59.06} & \textbf{67.62} & \textbf{63.05} & \textbf{61.31} & \textbf{65.36} & \textbf{66.07} & \textbf{62.90}               \\ \bottomrule
\end{tabular}
\caption{Cross-lingual zero-shot F1 accuracy on FUNSD and XFUND (fine-tuning on FUNSD, testing on X). Please note that only LiLT and LDM are pre-trained using pure English document data.}
\label{tab:zero-shot}
\vspace{-12pt}
\end{table*}

\subsection{The Language-Independent Pre-training}

Based on the analysis above, we conclude that a VAE-based text editing model effectively decouples language bias from images and that language-independent data can enhance cross-lingual capabilities. Inspired by this, we propose a straightforward language-independent paradigm, LDP, which utilizes monolingual corpus for pre-training.

Following ESP, we use DocBank and RVL-CDIP as our pre-training datasets and generate pseudo-labels. Subsequently, we use AnyText to decouple language bias from the original images by specifying the target region as $[0,0,1,1]$ and providing an empty prompt ``\_". In summary, we primarily replace the original image with a language-decoupled version while maintaining all other settings consistent with previous work. We pre-train our model using the language-independent data. In the fine-tuning, we maintain the original image to evaluate downstream performance.

\section{Model}

As illustrated in Figure \ref{fig: architecture}(a), our model is based on SAM, with the mask prediction head replaced by an MLP layer. We adhere to SAM's processing and image encoding procedure. 
We introduce \textbf{MTIM} (\textbf{M}ulti-\textbf{T}oken \textbf{I}nformation \textbf{M}erging) to integrate information from multiple bounding boxes within a single image. 
Given an image $I$, we first get all bounding box $b_n$ and their corresponding text $t_n$ from OCR information.
In the pre-training phase, only visual modality $I$ and layout modality $b_n$ are inputted to the model, and the feature from MTIM-augmented SAM is directly applied for pre-training.
After pre-training, we apply \textbf{L}anguage \textbf{K}nowledge \textbf{I}nserting (\textbf{LKI}) to enhance the model's text modality $t_n$ for downstream tasks.

\subsection{Multi-Token Information Merging (MTIM)}

When two text blocks share similar features, like the same background color, they are more likely to be classified under the same category. However, SAM handles different bounding boxes within a single image independently, limiting the model's ability to infer information from adjacent bounding boxes. To address this issue, we introduce MTIM.

As illustrated in Figure \ref{fig: architecture}(b), for each bounding box $b_n$, we get the multi-modality tokens $F_{nk}^{\text{SAM}}$ from the SAM decoder layer, where $k \in [0, K]$ and $K$ is the length pre-defined by SAM. MTIM module takes these tokens as input, and concatenates them to form a single feature vector:
\begin{equation}
F_{n}^{\text{Merge}} = CONCATE_{k=1}^K(F_{nk}^{\text{SAM}})
\end{equation}

\noindent All $F^{\text{Merge}}$ features within an image are then serialized into a sequence and passed through a self-attention layer followed by an MLP layer.
After interacting with different bounding box information, all vectors are resized to their original dimensions before being fed into the next layer. We use the final layer's MTIM feature $F_{n}^{\text{Final}}$ for pre-training.

\subsection{Language Knowledge Inserting (LKI)}

The text modality is excluded during pre-training to decouple language bias and enhance generalization, while in the fine-tuning stage, LKI is introduced to improve language-specific accuracy by incorporating language knowledge.

In detail, for text sequence $t_n$ from OCR information, we generate text embedding with a pre-trained, frozen multilingual embedding model, Sentence BERT \cite{Sentence-BERT,Multilingual-Sentence-BERT}. Sentence BERT is designed to map the text sequence to a fixed-size vector, and text with similar meanings will be encoded into nearby feature vectors. Then the text embedding is transformed into the same vector space with the final layer's MTIM feature, and fused for downstream tasks:
\begin{equation}
    F_n^{\text{Downstream}} = F_{n}^{\text{Final}} + Linear(BERT(t_n))
\end{equation}

\noindent In the downstream task, we add an MLP head to classify each bounding box to the target entity type.
A cross-entropy loss is adopted to end-to-end train the whole model.

\section{Experiments}

\begin{table*}[]
\centering
\begin{tabular}{lcccccccccc}
\toprule
\multirow{2}{*}{Model} & FUNSD & \multicolumn{7}{c}{XFUND}                             & \multicolumn{2}{c}{Avg.}  \\ \cmidrule(l){2-2} \cmidrule(l){3-9} \cmidrule(l){10-11}
                       & EN    & ZH    & JA    & ES    & FR    & IT    & DE    & PT    & All   & w/o EN                  \\ \midrule
XLM-RoBERTa            & 66.70 & 87.74 & 77.61 & 61.05 & 67.43 & 66.87 & 68.14 & 68.18 & 70.47 & 71.01               \\
InfoXLM                & 68.52 & 88.68 & 78.65 & 62.30 & 70.15 & 67.51 & 70.63 & 70.08 & 72.07 & 72.58              \\ \midrule
LayoutXLM              & 79.40 & 89.24 & 79.21 & 75.50 & 79.02 & 80.82 & 82.22 & 79.03 & 80.56 & 80.72               \\
LiLT                   & 84.15 & 89.38 & 79.64 & 79.11 & 79.53 & 83.76 & 82.31 & 82.20 & 82.51 & 82.28               \\ 
ESP                    & \textbf{91.10} & 90.30 & 81.10 & 85.40 & \textbf{90.50} & \textbf{88.90} & 87.20 & 87.50 & 87.30 & 86.76               \\ \midrule
\textbf{LDM(Ours)}                    & 88.23 & \textbf{91.08} & \textbf{82.62} & \textbf{86.60} & 89.79 & 88.53 & \textbf{89.78} & \textbf{89.10} & \textbf{88.21} & \textbf{88.21}              \\ \bottomrule
\end{tabular}
\vspace{-2pt}
\caption{Language-specific fine-tuning F1 accuracy on FUNSD and XFUND (fine-tuning on X, testing on X).}
\label{tab:specific}
\end{table*}

\begin{table*}[]
\centering
\begin{tabular}{lcccccccccc}
\toprule
\multirow{2}{*}{Model} & FUNSD & \multicolumn{7}{c}{XFUND}                             & \multicolumn{2}{c}{Avg.}  \\ \cmidrule(l){2-2} \cmidrule(l){3-9} \cmidrule(l){10-11}
                       & EN    & ZH    & JA    & ES    & FR    & IT    & DE    & PT    & All   & w/o EN                  \\ \midrule
XLM-RoBERTa            & 66.33 & 88.30 & 77.86 & 62.23 & 70.35 & 68.14 & 71.46 & 67.35 & 71.49 & 72.23               \\
InfoXLM                & 65.38 & 87.41 & 78.55 & 59.79 & 70.57 & 68.26 & 70.55 & 67.96 & 71.06 & 71.87               \\ \midrule
LayoutXLM              & 79.24 & 89.73 & 79.64 & 77.98 & 81.73 & 82.10 & 83.22 & 82.41 & 82.01 & 82.41               \\
LiLT                   & 85.74 & 90.47 & 80.88 & 83.40 & 85.77 & 87.92 & 87.69 & 84.93 & 85.85 & 85.87              \\ \midrule
\textbf{LDM(Ours)}                    & \textbf{89.78} & \textbf{91.86} & \textbf{83.67} & \textbf{88.02} & \textbf{91.16} & \textbf{89.95} & \textbf{90.83} & \textbf{90.34} & \textbf{89.45} & \textbf{89.40}              \\ \bottomrule
\end{tabular}
\vspace{-2pt}
\caption{Multitask fine-tuning F1 accuracy on FUNSD and XFUND (fine-tuning on 8 languages all, testing on X).}
\label{tab:multi-task}
\vspace{-12pt}
\end{table*}

\subsection{Datasets}

\noindent\textbf{Pre-Training.} Following StrucTexT and ESP, We use DocBank \cite{DocBank} and RVL-CDIP \cite{RVL-CDIP} to pre-train our model.
DocBank is a fine-grained document layout analysis dataset consisting of 500K images with corresponding OCR annotations. DocBank provides word-level and paragraph-level annotations. We heuristically generate OCR annotations by merging words in a line with overlapping y-coordinates and nearby x-coordinates within the same paragraph.
RVL-CDIP is a document image classification dataset consisting of 400K images categorized into 16 classes. Since RVL-CDIP lacks OCR annotations, we use PaddleOCR to extract the necessary OCR information.
The pseudo labels are generated using the algorithm described in ESP. We use AnyText to generate language-independent images. The ``decouple resolution" is set to 1024. The pseudo labels and language-independent images are used to pre-train our model, only the EE pre-training in ESP is applied.

\noindent\textbf{Fine-Tuning.} We evaluate the performance of LDM on both multilingual and monolingual datasets, with a primary focus on the Entity Extraction task.
\textbf{FUNSD} \cite{FUNSD} is a well-annotated English dataset for form understanding, containing 149 training examples and 50 testing samples. The task involves classifying semantic entities such as questions, answers, headers, and others.
\textbf{XFUND} \cite{XFUND} is a multilingual extension of FUNSD, including form understanding samples in seven non-English languages (Chinese, Japanese, Spanish, French, Italian, German, and Portuguese). It follows the same task definition as FUNSD, with 149 training samples and 50 testing samples for each language.
\textbf{SIBR} \cite{ESP} is a bilingual dataset (English and Chinese) characterized by diverse appearances and rich structures. It includes 600 training samples and 400 testing samples, following the same task definition as FUNSD.
\textbf{CORD} \cite{CORD} is an English dataset consisting of camera-captured receipts, featuring more detailed classifications such as menu name, price, quantity, \textit{etc}. It contains 800 training samples, 100 validation samples, and 100 testing samples.
All fine-tuning datasets provide fine-grained OCR annotations, and we directly use them as our OCR information. All fine-tuning datasets apply F1 as the evaluation metric.

\subsection{Implementation Details}

The LDM model is built using the PyTorch framework and the Hugging Face Transformers library. We adhere to all preprocessing steps and pre-trained parameters from $\text{SAM}_\text{BASE}$, except for the prediction head. All other parameters are randomly initialized.
The LDM model is trained using the Adam optimizer with a learning rate of 2e-4. The learning rate is linearly warmed up for the first 10\% of steps, followed by cosine decay. The training batch size is set to 32. The LDM model is pre-trained for 10 epochs and fine-tuned for 2000 steps using 8 NVIDIA A6000 48GB GPUs. During pre-training, the number of bounding boxes is truncated to 512, while in fine-tuning, all bounding boxes are retained.

\begin{table}[]
\centering
\begin{tabular}{lccc}
\toprule
\textbf{Model}   &  \textbf{Precision}   &  \textbf{Recall}   &  \textbf{F1}  \\ \midrule
TRIE              &  -                   &  -                    &  85.62 \\
LayoutXLM         &  -                   &  -                    &  94.72 \\ 
ESP               &  -                   &  -                    &  95.27 \\ \midrule
\textbf{LDM(Ours)}         &  96.07                &  95.14                &  \textbf{95.60} \\ \bottomrule
\end{tabular}
\caption{Performance on SIBR.}
\vspace{-8pt}
\label{tab:SIBR}
\end{table}

\subsection{Comparison with SOTA Methods}

\noindent\textbf{Cross-Lingual Zero-Shot Generalization.} The results are presented in Table \ref{tab:zero-shot}. This setup requires the model to fine-tune on English (FUNSD) and then generalize to non-English scenarios (XFUND).
LDM demonstrates state-of-the-art performance among multilingual VIE models such as LiLT and LayoutXLM. 
Specifically, LDM exhibits superior generalization on non-English subsets. For example, in FUNSD, LDM outperforms LiLT by 4.08\% (88.23\% vs. 84.15\%), while in XFUND, LDM achieves a higher margin of 5.66\% (62.9\% vs. 57.24\%). This further underscores the excellent cross-lingual generalization of our proposed LDP training paradigm and LDM model.

\noindent\textbf{Language-Specific Fine-Tuning.} In this experimental setup, all language subsets are fine-tuned and evaluated individually.
As shown in Table \ref{tab:specific}, LDM still outperforms all text-dominated methods. LDM and ESP are both pre-trained on DocBank and RVL-CDIP, using visual images as the primary input, with pseudo-labels generated by the same algorithm. The key difference lies in the use of language-independent images.
ESP introduces English knowledge during pre-training, resulting in better performance on English datasets. However, this approach also leads to overfitting, our language-independent pre-training allows our LDM model to maintain superior generalization on non-English datasets (88.21\% for LDM vs. 86.76\% for ESP).

\noindent\textbf{Multitask Fine-Tuning.} As illustrated in Table \ref{tab:multi-task}, when all language data are fine-tuned together, LDM continues to demonstrate SOTA performance.
Notably, in this setting, all multi-modality models achieve better performance compared to language-specific fine-tuning, whereas the accuracy of pure NLP models like InfoXLM decreases slightly.
This result further suggests that document images in different languages share commonalities in layout and visual modalities.

\begin{table}[]
\centering
\begin{tabular}{lccc}
\toprule
\textbf{Model}   &  \textbf{Precision}   &  \textbf{Recall}   &  \textbf{F1}  \\ \midrule
LayoutLM         &  75.97                &  81.55                &  78.66 \\
BROS             &  80.56                &  81.88                &  81.21 \\
LayoutLMv2       &  80.29                &  85.39                &  82.76 \\
StrucTexT        &  85.68                &  80.97                &  83.09 \\
LayoutLMv3       &  -                    &  -                    &  90.29 \\
UDOP             &  -                    &  -                    &  91.62 \\ \midrule
LayoutXLM        &  79.13                &  81.58                &  80.34 \\
LiLT             &  \underline{\textit{84.67}}                &  \underline{\textit{87.09}}                &  85.86 \\
ESP              &  -                    &  -                    &  \textbf{91.12} \\ \midrule
\textbf{LDM(Ours)}         &  \textbf{88.45}                &  \textbf{88.01}                &  \underline{\textit{88.23}} \\ \bottomrule
\end{tabular}
\caption{Performance on FUNSD. The best multilingual model is in \textbf{bold}, and the second is in \textit{\underline{italics}}.}
\vspace{-4pt}
\label{tab:FUNSD}
\end{table}

\noindent\textbf{Bilingual/English Dataset.} We conducted experiments on a broader range of VIE datasets to better evaluate the performance of LDM.
As shown in Table \ref{tab:SIBR},  LDM continues to demonstrate state-of-the-art performance on the bilingual (English and Chinese) SIBR dataset, despite the presence of challenging scenarios such as image blur and printing shift.
Table \ref{tab:FUNSD} and Table \ref{tab:CORD} illustrate the performance on English-only datasets. Although pre-trained for multilingual scenarios, LDM still achieve better accuracy than most English-specific models, such as LayoutLMv2. When compared to multilingual text-based models like LayoutXLM and LiLT, LDM consistently outperforms them.

\begin{table}[]
\centering
\begin{tabular}{lccc}
\toprule
\textbf{Model}   &  \textbf{Precision}   &  \textbf{Recall}   &  \textbf{F1}  \\ \midrule
LayoutLM         &  94.37                &  95.08                &  94.72 \\
BROS             &  95.58                &  95.14                &  95.36 \\
TILT             &  -                    &  -                    &  95.11 \\
LayoutLMv2       &  94.53                &  95.39                &  94.95 \\
DocFormer        &  96.52                &  96.14                &  96.33 \\
LayoutLMv3       &  -                    &  -                    &  96.56 \\
UDOP             &  -                    &  -                    &  97.58 \\ \midrule
LayoutXLM        &  94.56                &  95.06                &  94.81 \\
LiLT             &  \underline{\textit{95.74}}                &  \textbf{95.81}                &  \underline{\textit{95.77}} \\
ESP              &  -                    &  -                    &  95.65 \\ \midrule
\textbf{LDM(Ours)}         &  \textbf{95.97}                &  \underline{\textit{95.64}}                &  \textbf{95.80} \\ \bottomrule
\end{tabular}
\caption{Performance on CORD. The best multilingual model is in \textbf{bold}, and the second is in \textit{\underline{italics}}.}
\label{tab:CORD}
\vspace{-12pt}
\end{table}

\vspace{-5pt}
\subsection{Ablation Study}

\begin{table}[]
\centering
\begin{tabular}{ccccc}
\toprule
\#  & Decouple Resolution   & MTIM                         & FUNSD & XFUND \\ \midrule
1(a) & \multirow{2}{*}{N/A}  & \usym{2717} & 84.55 & 57.25 \\
1(b) &                       & \usym{2713} & \textbf{86.24} & 57.65 \\ \midrule
2(a) & \multirow{2}{*}{2048} & \usym{2717} & 83.87 & 59.18 \\
2(b) &                       & \usym{2713} & \underline{\textit{85.91}} & \underline{\textit{59.82}} \\ \midrule
3(a) & \multirow{2}{*}{1536} & \usym{2717} & 83.10 & 59.71 \\
3(b) &                       & \usym{2713} & 85.53 & \underline{\textit{60.25}} \\ \midrule
4(a) & \multirow{2}{*}{1024} & \usym{2717} & 83.01 & 59.46 \\
4(b) &                       & \usym{2713} & \underline{\textit{85.54}} & \textbf{60.68} \\ \midrule
5(a) & \multirow{2}{*}{768}  & \usym{2717} & 81.82 & 57.97 \\
5(b) &                       & \usym{2713} & 83.10 & 59.95 \\ \midrule
6(a) & \multirow{2}{*}{-}  & \usym{2717} & 80.03 & 56.38 \\
6(b) &                       & \usym{2713} & 54.58 & 44.76 \\ \bottomrule
\end{tabular}
\caption{Ablation study for pre-training. LDM is fine-tuned on FUNSD, and zero-shot evaluated on XFUND. ``N/A" means language decoupling is not applied, and the model is pre-trained using original images. ``-" means LDM is not pre-trained and only initialized from the SAM's parameters.}
\label{tab:pretrain}
\end{table}

\noindent\textbf{Effect of Language-Independent Pre-training Data.}
We conduct ablation studies on FUNSD and XFUND to evaluate the impact of introducing language-independent data into pre-training. Pre-training is limited to a single epoch due to the time-intensive nature of these experiments.
As shown in Table \ref{tab:pretrain}, as the ``decouple resolution" decreases, namely more language bias is decoupled, the cross-lingual generalization (XFUND) gradually improves, while the English accuracy only decreases slightly (See \#1(b), \#2(b), \#3(b) and \#4(b)).
The cross-lingual performance only drops when the ``decouple resolution" becomes too low (See \#4(b) and \#5(b)), which we attribute to extreme information loss at low resolutions.
In our final experiments, we set ``decouple resolution" to 1024, as it offers the best trade-off in all settings.

\noindent\textbf{Effect of MTIM.} When applying pre-training, MTIM consistently introduces performance improvements, verifying the effectiveness of integrating different bounding box information. In the absence of pre-training, we attribute the performance decrease to unsuitable parameters, as all other parameters are inherited from SAM and MTIM is randomly initialized, which can disrupt information interaction if there is insufficient training data.

\begin{table}[]
\centering
\begin{tabular}{ccccc}
\toprule
\multirow{2}{*}{\#} & \multicolumn{2}{c}{LKI} & \multirow{2}{*}{FUNSD} & \multirow{2}{*}{XFUND} \\ \cmidrule(l){2-3}
                    & Decoder   & Classifier  &                        &                        \\ \midrule
1                   &           &             & 86.95                       & 61.36                       \\
2                   & \usym{2713}          &             & 87.37                       & 61.51                       \\
3                   &           &  \usym{2713}           & \textbf{88.23}                       & 62.90                       \\
4                   & \usym{2713}          & \usym{2713}            & 88.07                      & \textbf{62.99}                       \\ \bottomrule
\end{tabular}
\caption{Ablation studies for LKI module.}
\label{tab: LKI}
\vspace{-12pt}
\end{table}

\noindent\textbf{Effect of LKI.} 
We conduct experiments to evaluate the effectiveness of incorporating language knowledge in downstream tasks.
In addition to fusing language knowledge at the classification head, we also attempted to fuse it into the first decoder layer. 
To avoid mismatch parameters like in MTIM ablation studies, we initialize the fuse layer in decoder with zero.
As shown in Table \ref{tab: LKI}, incorporating language knowledge consistently improves performance, as seen in \#1, \#2, and \#3. Fusing language knowledge at the classification head yields the highest improvement.
Comparing \#3 and \#4, jointly inserting language knowledge into both the decoder layer and classification head does not result in a significant improvement.
Therefore, in our final experiment, we choose to add LKI to the classification head.

\vspace{-5pt}
\section{Conclusion}

In this paper, we conduct systematic experiments to decouple language bias from document images. We propose a multilingual pre-training paradigm LDP to transfer from monolingual data to multilingual ones. Our experimental results on downstream benchmarks validate that LDP can significantly improve the cross-lingual generalization in visual document understanding.
For future research, we will try to dig deeper to the invariance among different languages and try to integrate text modality into pre-training.


\section{Acknowledgments}
Supported by the National Natural Science Foundation of China (Grant NO 62376266 and 62406318), and by the Key Research Program of Frontier Sciences, CAS (Grant NO ZDBS-LY-7024).

\bibliography{aaai25}

\begin{thebibliography}{46}
\providecommand{\natexlab}[1]{#1}

\bibitem[{Appalaraju et~al.(2024)Appalaraju, Tang, Dong, Sankaran, Zhou, and Manmatha}]{DocFormerv2}
Appalaraju, S.; Tang, P.; Dong, Q.; Sankaran, N.; Zhou, Y.; and Manmatha, R. 2024.
\newblock DocFormerv2: Local Features for Document Understanding.
\newblock In \emph{Proceedings of the AAAI Conference on Artificial Intelligence (AAAI)}.

\bibitem[{Chen et~al.(2023)Chen, Xu, Gu, Lan, Zheng, Li, Meng, Zhu, and Wang}]{VAE-Text2}
Chen, H.; Xu, Z.; Gu, Z.; Lan, J.; Zheng, X.; Li, Y.; Meng, C.; Zhu, H.; and Wang, W. 2023.
\newblock DiffUTE: Universal Text Editing Diffusion Model.
\newblock In \emph{Advances in Neural Information Processing Systems (NeurIPS)}.

\bibitem[{Chen et~al.(2020)Chen, Wang, Zhou, Yang, Yang, and Wang}]{TextDetection}
Chen, Y.; Wang, W.; Zhou, Y.; Yang, F.; Yang, D.; and Wang, W. 2020.
\newblock Self-Training for Domain Adaptive Scene Text Detection.
\newblock In \emph{International Conference on Pattern Recognition, (ICPR)}.

\bibitem[{Cheng et~al.(2022)Cheng, Zhang, Li, Liang, Xu, Li, Pu, Niu, and Wu}]{TRIE++}
Cheng, Z.; Zhang, P.; Li, C.; Liang, Q.; Xu, Y.; Li, P.; Pu, S.; Niu, Y.; and Wu, F. 2022.
\newblock TRIE++: Towards End-to-End Information Extraction from Visually Rich Documents.
\newblock arXiv:2207.06744.

\bibitem[{Deng et~al.(2009)Deng, Dong, Socher, Li, Li, and Fei-Fei}]{ImageNet}
Deng, J.; Dong, W.; Socher, R.; Li, L.-J.; Li, K.; and Fei-Fei, L. 2009.
\newblock ImageNet: A Large-Scale Hierarchical Image Database.
\newblock In \emph{Proceedings of the IEEE Conference on Computer Vision and Pattern Recognition (CVPR)}.

\bibitem[{Devlin et~al.(2019)Devlin, Chang, Lee, and Toutanova}]{BERT}
Devlin, J.; Chang, M.-W.; Lee, K.; and Toutanova, K. 2019.
\newblock BERT: Pre-Training of Deep Bidirectional Transformers for Language Understanding.
\newblock In \emph{Proceedings of the Conference of the North American Chapter of the Association for Computational Linguistics: Human Language Technologies (NAACL-HLT)}.

\bibitem[{Fujitake(2024)}]{LayoutLLM}
Fujitake, M. 2024.
\newblock LayoutLLM: Large Language Model Instruction Tuning for Visually Rich Document Understanding.
\newblock In \emph{Proceedings of the Joint International Conference on Computational Linguistics and Language Resources and Evaluation (LREC/COLING)}.

\bibitem[{Gu et~al.(2022)Gu, Meng, Wang, Lan, Wang, Gu, and Zhang}]{XYLayoutLM}
Gu, Z.; Meng, C.; Wang, K.; Lan, J.; Wang, W.; Gu, M.; and Zhang, L. 2022.
\newblock XYLayoutLM: Towards Layout-Aware Multimodal Networks for Visually-Rich Document Understanding.
\newblock In \emph{Proceedings of the IEEE/CVF Conference on Computer Vision and Pattern Recognition (CVPR)}.

\bibitem[{Harley, Ufkes, and Derpanis(2015)}]{RVL-CDIP}
Harley, A.~W.; Ufkes, A.; and Derpanis, K.~G. 2015.
\newblock Evaluation of Deep Convolutional Nets for Document Image Classification and Retrieval.
\newblock In \emph{Proceedings of the International Conference on Document Analysis and Recognition (ICDAR)}.

\bibitem[{Hu et~al.(2024)Hu, Xu, Ye, Yan, Zhang, Zhang, Li, Zhang, Jin, Huang, and Zhou}]{docowl-1.5}
Hu, A.; Xu, H.; Ye, J.; Yan, M.; Zhang, L.; Zhang, B.; Li, C.; Zhang, J.; Jin, Q.; Huang, F.; and Zhou, J. 2024.
\newblock mPLUG-DocOwl 1.5: Unified Structure Learning for OCR-free Document Understanding.
\newblock arXiv:2403.12895.

\bibitem[{Huang et~al.(2022)Huang, Lv, Cui, Lu, and Wei}]{LayoutLMv3}
Huang, Y.; Lv, T.; Cui, L.; Lu, Y.; and Wei, F. 2022.
\newblock LayoutLMv3: Pre-Training for Document AI with Unified Text and Image Masking.
\newblock In \emph{Proceedings of the ACM International Conference on Multimedia (MM)}.

\bibitem[{Jaume, Ekenel, and Thiran(2019)}]{FUNSD}
Jaume, G.; Ekenel, H.~K.; and Thiran, J.-P. 2019.
\newblock FUNSD: A Dataset for Form Understanding in Noisy Scanned Documents.
\newblock In \emph{Proceedings of the International Conference on Document Analysis and Recognition Workshops (ICDARW)}.

\bibitem[{Kim et~al.(2022)Kim, Hong, Yim, Nam, Park, Yim, Hwang, Yun, Han, and Park}]{Donut}
Kim, G.; Hong, T.; Yim, M.; Nam, J.; Park, J.; Yim, J.; Hwang, W.; Yun, S.; Han, D.; and Park, S. 2022.
\newblock OCR-Free Document Understanding Transformer.
\newblock In \emph{Proceedings of the European Conference on Computer Vision (ECCV)}.

\bibitem[{Kingma and Welling(2014)}]{VAE}
Kingma, D.~P.; and Welling, M. 2014.
\newblock Auto-Encoding Variational Bayes.
\newblock In \emph{Proceedings of the International Conference on Learning Representations (ICLR)}.

\bibitem[{Kirillov et~al.(2023)Kirillov, Mintun, Ravi, Mao, Rolland, Gustafson, Xiao, Whitehead, Berg, Lo et~al.}]{SAM}
Kirillov, A.; Mintun, E.; Ravi, N.; Mao, H.; Rolland, C.; Gustafson, L.; Xiao, T.; Whitehead, S.; Berg, A.~C.; Lo, W.-Y.; et~al. 2023.
\newblock Segment Anything.
\newblock In \emph{Proceedings of the IEEE/CVF International Conference on Computer Vision (ICCV)}.

\bibitem[{Lewis et~al.(2006)Lewis, Agam, Argamon, Frieder, Grossman, and Heard}]{IIT-CDIP}
Lewis, D.~D.; Agam, G.; Argamon, S.; Frieder, O.; Grossman, D.~A.; and Heard, J. 2006.
\newblock Building a Test Collection for Complex Document Information Processing.
\newblock In \emph{Proceedings of the Annual International ACM SIGIR Conference on Research and Development in Information Retrieval (SIGIR)}.

\bibitem[{Li et~al.(2020)Li, Xu, Cui, Huang, Wei, Li, and Zhou}]{DocBank}
Li, M.; Xu, Y.; Cui, L.; Huang, S.; Wei, F.; Li, Z.; and Zhou, M. 2020.
\newblock DocBank: A Benchmark Dataset for Document Layout Analysis.
\newblock In \emph{Proceedings of the International Conference on Computational Linguistics (COLING)}.

\bibitem[{Li et~al.(2021)Li, Qian, Yu, Qin, Zhang, Liu, Yao, Han, Liu, and Ding}]{StrucTexT}
Li, Y.; Qian, Y.; Yu, Y.; Qin, X.; Zhang, C.; Liu, Y.; Yao, K.; Han, J.; Liu, J.; and Ding, E. 2021.
\newblock StrucTexT: Structured Text Understanding with Multi-Modal Transformers.
\newblock In \emph{Proceedings of the ACM Multimedia Conference (MM)}.

\bibitem[{Li et~al.(2024)Li, Shu, Zeng, Yang, and Zhou}]{VAE-Text1}
Li, Z.; Shu, Y.; Zeng, W.; Yang, D.; and Zhou, Y. 2024.
\newblock First Creating Backgrounds Then Rendering Texts: A New Paradigm for Visual Text Blending.
\newblock In \emph{European Conference on Artificial Intelligence (ECAI)}.

\bibitem[{Liu et~al.(2023)Liu, Li, Wu, and Lee}]{LLaVA}
Liu, H.; Li, C.; Wu, Q.; and Lee, Y.~J. 2023.
\newblock Visual Instruction Tuning.
\newblock In \emph{Advances in Neural Information Processing Systems (NeurIPS)}.

\bibitem[{Liu et~al.(2024)Liu, Yang, Liu, Li, Ma, Zhang, and Bai}]{textmonkey}
Liu, Y.; Yang, B.; Liu, Q.; Li, Z.; Ma, Z.; Zhang, S.; and Bai, X. 2024.
\newblock TextMonkey: An OCR-Free Large Multimodal Model for Understanding Document.
\newblock arXiv:2403.04473.

\bibitem[{Park et~al.(2019)Park, Shin, Lee, Lee, Surh, Seo, and Lee}]{CORD}
Park, S.; Shin, S.; Lee, B.; Lee, J.; Surh, J.; Seo, M.; and Lee, H. 2019.
\newblock CORD: A Consolidated Receipt Dataset for Post-OCR Parsing.
\newblock In \emph{Proceedings of the Workshop on Document Intelligence at NeurIPS}.

\bibitem[{Qiao et~al.(2021)Qiao, Zhou, Wei, Wang, Zhang, Jiang, Wang, and Wang}]{PIMNet}
Qiao, Z.; Zhou, Y.; Wei, J.; Wang, W.; Zhang, Y.; Jiang, N.; Wang, H.; and Wang, W. 2021.
\newblock PIMNet: {A} Parallel, Iterative and Mimicking Network for Scene Text Recognition.
\newblock In \emph{Proceedings of the ACM International Conference on Multimedia (MM)}.

\bibitem[{Qiao et~al.(2020)Qiao, Zhou, Yang, Zhou, and Wang}]{SEED}
Qiao, Z.; Zhou, Y.; Yang, D.; Zhou, Y.; and Wang, W. 2020.
\newblock {SEED:} Semantics Enhanced Encoder-Decoder Framework for Scene Text Recognition.
\newblock In \emph{Proceedings of the IEEE Conference on Computer Vision and Pattern Recognition (CVPR)}.

\bibitem[{Raffel et~al.(2020)Raffel, Shazeer, Roberts, Lee, Narang, Matena, Zhou, Li, and Liu}]{T5}
Raffel, C.; Shazeer, N.; Roberts, A.; Lee, K.; Narang, S.; Matena, M.; Zhou, Y.; Li, W.; and Liu, P.~J. 2020.
\newblock Exploring the Limits of Transfer Learning with a Unified Text-to-Text Transformer.
\newblock \emph{Journal of Machine Learning Research (JMLR)}.

\bibitem[{Reimers and Gurevych(2019)}]{Sentence-BERT}
Reimers, N.; and Gurevych, I. 2019.
\newblock Sentence-BERT: Sentence Embeddings Using Siamese BERT-Networks.
\newblock In \emph{Proceedings of the Conference on Empirical Methods in Natural Language Processing and the International Joint Conference on Natural Language Processing (EMNLP-IJCNLP)}.

\bibitem[{Reimers and Gurevych(2020)}]{Multilingual-Sentence-BERT}
Reimers, N.; and Gurevych, I. 2020.
\newblock Making Monolingual Sentence Embeddings Multilingual using Knowledge Distillation.
\newblock In \emph{Proceedings of the Conference on Empirical Methods in Natural Language Processing (EMNLP)}.

\bibitem[{Shen et~al.(2023)Shen, Gao, Wei, Qiao, Zhou, Li, and Cheng}]{DRCC}
Shen, H.; Gao, X.; Wei, J.; Qiao, L.; Zhou, Y.; Li, Q.; and Cheng, Z. 2023.
\newblock Divide Rows and Conquer Cells: Towards Structure Recognition for Large Tables.
\newblock In \emph{Proceedings of the Thirty-Second International Joint Conference on Artificial Intelligence, (IJCAI)}.

\bibitem[{Shu et~al.(2024)Shu, Zeng, Li, Zhao, and Zhou}]{TextSurvey}
Shu, Y.; Zeng, W.; Li, Z.; Zhao, F.; and Zhou, Y. 2024.
\newblock Visual Text Meets Low-level Vision: {A} Comprehensive Survey on Visual Text Processing.
\newblock arXiv:2402.03082.

\bibitem[{Tang et~al.(2023)Tang, Yang, Wang, Fang, Liu, Zhu, Zeng, Zhang, and Bansal}]{UDOP}
Tang, Z.; Yang, Z.; Wang, G.; Fang, Y.; Liu, Y.; Zhu, C.; Zeng, M.; Zhang, C.; and Bansal, M. 2023.
\newblock Unifying Vision, Text, and Layout for Universal Document Processing.
\newblock In \emph{Proceedings of the IEEE/CVF Conference on Computer Vision and Pattern Recognition (CVPR)}.

\bibitem[{Touvron et~al.(2023)Touvron, Lavril, Izacard, Martinet, Lachaux, Lacroix, Rozière, Goyal, Hambro, Azhar, Rodriguez, Joulin, Grave, and Lample}]{LLaMA}
Touvron, H.; Lavril, T.; Izacard, G.; Martinet, X.; Lachaux, M.-A.; Lacroix, T.; Rozière, B.; Goyal, N.; Hambro, E.; Azhar, F.; Rodriguez, A.; Joulin, A.; Grave, E.; and Lample, G. 2023.
\newblock LLaMA: Open and Efficient Foundation Language Models.
\newblock arXiv:2302.13971.

\bibitem[{Tuo et~al.(2024)Tuo, Xiang, He, Geng, and Xie}]{AnyText}
Tuo, Y.; Xiang, W.; He, J.-Y.; Geng, Y.; and Xie, X. 2024.
\newblock AnyText: Multilingual Visual Text Generation and Editing.
\newblock In \emph{Proceedings of the International Conference on Learning Representations (ICLR)}.

\bibitem[{Wang, Jin, and Ding(2022)}]{LiLT}
Wang, J.; Jin, L.; and Ding, K. 2022.
\newblock LiLT: A Simple yet Effective Language-Independent Layout Transformer for Structured Document Understanding.
\newblock In \emph{Proceedings of the Annual Meeting of the Association for Computational Linguistics (ACL)}.

\bibitem[{Wei et~al.(2025)Wei, Kong, Chen, Zhao, Ge, Yang, Sun, Han, and Zhang}]{Vary}
Wei, H.; Kong, L.; Chen, J.; Zhao, L.; Ge, Z.; Yang, J.; Sun, J.; Han, C.; and Zhang, X. 2025.
\newblock Vary: Scaling up the vision vocabulary for large vision-language model.
\newblock In \emph{European Conference on Computer Vision (ECCV)}.

\bibitem[{Xu et~al.(2020)Xu, Li, Cui, Huang, Wei, and Zhou}]{LayoutLM}
Xu, Y.; Li, M.; Cui, L.; Huang, S.; Wei, F.; and Zhou, M. 2020.
\newblock LayoutLM: Pre-Training of Text and Layout for Document Image Understanding.
\newblock In \emph{Proceedings of the ACM SIGKDD Conference on Knowledge Discovery and Data Mining (KDD)}.

\bibitem[{Xu et~al.(2021{\natexlab{a}})Xu, Lv, Cui, Wang, Lu, Florencio, Zhang, and Wei}]{layoutxlm}
Xu, Y.; Lv, T.; Cui, L.; Wang, G.; Lu, Y.; Florencio, D.; Zhang, C.; and Wei, F. 2021{\natexlab{a}}.
\newblock LayoutXLM: Multimodal Pre-training for Multilingual Visually-rich Document Understanding.
\newblock arXiv:2104.08836.

\bibitem[{Xu et~al.(2022)Xu, Lv, Cui, Wang, Lu, Florêncio, Zhang, and Wei}]{XFUND}
Xu, Y.; Lv, T.; Cui, L.; Wang, G.; Lu, Y.; Florêncio, D. A.~F.; Zhang, C.; and Wei, F. 2022.
\newblock XFUND: A Benchmark Dataset for Multilingual Visually Rich Form Understanding.
\newblock In \emph{Findings of the Association for Computational Linguistics (ACL)}.

\bibitem[{Xu et~al.(2021{\natexlab{b}})Xu, Xu, Lv, Cui, Wei, Wang, Lu, Florêncio, Zhang, Che, Zhang, and Zhou}]{LayoutLMv2}
Xu, Y.; Xu, Y.; Lv, T.; Cui, L.; Wei, F.; Wang, G.; Lu, Y.; Florêncio, D. A.~F.; Zhang, C.; Che, W.; Zhang, M.; and Zhou, L. 2021{\natexlab{b}}.
\newblock LayoutLMv2: Multi-Modal Pre-Training for Visually-Rich Document Understanding.
\newblock In \emph{Proceedings of the Annual Meeting of the Association for Computational Linguistics and the International Joint Conference on Natural Language Processing (ACL/IJCNLP)}.

\bibitem[{Yang et~al.(2023)Yang, Long, Wang, Song, Zhong, Cheng, Bai, and Yao}]{ESP}
Yang, Z.; Long, R.; Wang, P.; Song, S.; Zhong, H.; Cheng, W.; Bai, X.; and Yao, C. 2023.
\newblock Modeling Entities as Semantic Points for Visual Information Extraction in the Wild.
\newblock In \emph{Proceedings of the IEEE/CVF Conference on Computer Vision and Pattern Recognition (CVPR)}.

\bibitem[{Ye et~al.(2023{\natexlab{a}})Ye, Hu, Xu, Ye, Yan, Dan, Zhao, Xu, Li, Tian, Qi, Zhang, and Huang}]{mPLUG-DocOwl}
Ye, J.; Hu, A.; Xu, H.; Ye, Q.; Yan, M.; Dan, Y.; Zhao, C.; Xu, G.; Li, C.; Tian, J.; Qi, Q.; Zhang, J.; and Huang, F. 2023{\natexlab{a}}.
\newblock mPLUG-DocOwl: Modularized Multimodal Large Language Model for Document Understanding.
\newblock arXiv:2307.02499.

\bibitem[{Ye et~al.(2023{\natexlab{b}})Ye, Xu, Xu, Ye, Yan, Zhou, Wang, Hu, Shi, Shi, Li, Xu, Chen, Tian, Qi, Zhang, and Huang}]{mPLUG-Owl}
Ye, Q.; Xu, H.; Xu, G.; Ye, J.; Yan, M.; Zhou, Y.; Wang, J.; Hu, A.; Shi, P.; Shi, Y.; Li, C.; Xu, Y.; Chen, H.; Tian, J.; Qi, Q.; Zhang, J.; and Huang, F. 2023{\natexlab{b}}.
\newblock mPLUG-Owl: Modularization Empowers Large Language Models with Multimodality.
\newblock arXiv:2304.14178.

\bibitem[{Yu et~al.(2023)Yu, Li, Zhang, Zhang, Guo, Qin, Yao, Han, Ding, and Wang}]{StrucTexTv2}
Yu, Y.; Li, Y.; Zhang, C.; Zhang, X.; Guo, Z.; Qin, X.; Yao, K.; Han, J.; Ding, E.; and Wang, J. 2023.
\newblock StrucTexTv2: Masked Visual-Textual Prediction for Document Image Pre-Training.
\newblock In \emph{Proceedings of the International Conference on Learning Representations (ICLR)}.

\bibitem[{Zeng et~al.(2024{\natexlab{a}})Zeng, Zhang, Wei, Yang, Zhang, Gao, Qin, and Zhou}]{TextRetrival}
Zeng, G.; Zhang, Y.; Wei, J.; Yang, D.; Zhang, P.; Gao, Y.; Qin, X.; and Zhou, Y. 2024{\natexlab{a}}.
\newblock Focus, Distinguish, and Prompt: Unleashing {CLIP} for Efficient and Flexible Scene Text Retrieval.
\newblock In \emph{Proceedings of the ACM International Conference on Multimedia (MM)}.

\bibitem[{Zeng et~al.(2023)Zeng, Zhang, Zhou, Yang, Jiang, Zhao, Wang, and Yin}]{BeyongVQA}
Zeng, G.; Zhang, Y.; Zhou, Y.; Yang, X.; Jiang, N.; Zhao, G.; Wang, W.; and Yin, X. 2023.
\newblock Beyond {OCR} + {VQA:} Towards end-to-end reading and reasoning for robust and accurate textvqa.
\newblock \emph{Pattern Recognit.}

\bibitem[{Zeng et~al.(2024{\natexlab{b}})Zeng, Shu, Li, Yang, and Zhou}]{TextCtrl}
Zeng, W.; Shu, Y.; Li, Z.; Yang, D.; and Zhou, Y. 2024{\natexlab{b}}.
\newblock TextCtrl: Diffusion-based Scene Text Editing with Prior Guidance Control.
\newblock In \emph{Advances in Neural Information Processing Systems (NeurIPS)}.

\bibitem[{Zhang et~al.(2020)Zhang, Xu, Cheng, Pu, Lu, Qiao, Niu, and Wu}]{TRIE}
Zhang, P.; Xu, Y.; Cheng, Z.; Pu, S.; Lu, J.; Qiao, L.; Niu, Y.; and Wu, F. 2020.
\newblock TRIE: End-to-End Text Reading and Information Extraction for Document Understanding.
\newblock In \emph{Proceedings of the ACM International Conference on Multimedia (MM)}.

\end{thebibliography}

\end{document}